\documentclass[krantz1,ChapterTOCs]{krantz} %See documentation for other class options
\usepackage{amssymb}
\usepackage{amsmath}
\usepackage{graphicx}
\usepackage{subfigure}
\usepackage{makeidx}
\usepackage{multicol}
\usepackage[export]{adjustbox}
\usepackage[justification=centering]{caption}
\usepackage{chngcntr}
\counterwithout{figure}{section}
\usepackage{float}

\usepackage{multirow}
\usepackage{booktabs}
\usepackage{algpseudocode}
\usepackage{algorithm}
\usepackage{bbold}
\usepackage[nobiblatex]{xurl}

\begin{document}

%\frontmatter

%\title{Krantz Template} %This is a placeholder titlepage, it will not be final.
%\author{Yours Truly}
%%%\maketitle

%%%Placeholder for front matter

%\halftitle

%\booktitle

%\locpage

%\include{frontmatter/dedication}
%\cleardoublepage
%\setcounter{page}{7} %previous pages will be reserved for frontmatter to be added in later.
%\tableofcontents
%\include{frontmatter/foreword}
%\include{frontmatter/preface}
%%\listoffigures
%%\listoftables
% \include{frontmatter/contributor}
%\include{frontmatter/symbollist}

\mainmatter

\renewcommand{\sectionmark}[1]{\markboth {}{}}
\renewcommand{\thesection}{\arabic{section}}

%\part{This is What a Part Would Look Like}
%\input{chapter_contents}
% \chapterauthors{$^1$Shinji KAWAKURA, $^1$Masayuki HIRAFUJI, $^2$Ryosuke SHIBASAKI}{$^1$Osaka Metropolitan University, $^2$University of Tokyo}

%%%%%%%% here is the title of paper %%%%%%%%%%
\chapter{Contrastive Left-Right Wearable Sensors (IMUs) Consistency Matching for HAR}
%%%%%%%% here is the title of paper %%%%%%%%%%

%%%%%%%%%%% here is the authors %%%%%%%%%%%%%
%\authors{
%     Dominique Nshimyimana \footnote{nshimyim@rptu.de},
%     Vitor Fortes Rey \footnote{fortes@dfki.uni-kl.de},
%     Paul Lukowicz \footnote{Paul.Lukowicz@dfki.de}
%     } \\
%\authors{$^1$$^2$$^3$Department of Computer Science, Rheinland-Pfälzische Technische Universität Kaiserslautern-Landau (RPTU), 67663 Kaiserslautern, Germany \\ $^2$$^3$ German Research Center for Artificial Intelligence (DFKI), 67663 Kaiserslautern, Germany}
%%%%%%%%%%% here is the authors %%%%%%%%%%%%%

%%%%%%%%%%% here is the authors %%%%%%%%%%%%%
\begin{multicols}{2}

\contributor{Dominique Nshimyimana\footnote{nshimyim@rptu.de}}{DFKI and RPTU Kaiserslautern} {}\\

\contributor{Vitor Fortes Rey \footnote{fortes@dfki.uni-kl.de}}{DFKI and RPTU Kaiserslautern}{}\\

\contributor{Paul Lukowicz \footnote{Paul.Lukowicz@dfki.de}}{DFKI and RPTU Kaiserslautern}{}\\

\end{multicols}

%\authors{
%Anonymous Authors
%}

%\end{multicols}
%%%%%%%%%%% here is the authors %%%%%%%%%%%%%

%%%%%%%% abstract from here %%%%%%%%%%
\section*{Abstract}
Machine learning algorithms are improving rapidly, but annotating training data remains a bottleneck for many applications. In this paper, we show how real data can be used for self-supervised learning without any transformations by taking advantage of the symmetry present in the activities. Our approach involves contrastive matching of two different sensors (left and right wrist or leg-worn IMUs) to make representations of co-occurring sensor data more similar and those of non-co-occurring sensor data more different. We test our approach on the Opportunity and MM-Fit datasets. In MM-Fit we show significant improvement over the baseline supervised and self-supervised method SimCLR, while for Opportunity there is significant improvement over the supervised baseline and slight improvement when compared to SimCLR. Moreover, our method improves supervised baselines even when using only a small amount of the data for training. Future work should explore under which conditions our method is beneficial for human activity recognition systems and other related applications.

\section{Introduction}
\label{section:Introduction}

\begin{figure}
  \centering
  \includegraphics[page=1,width=\textwidth]{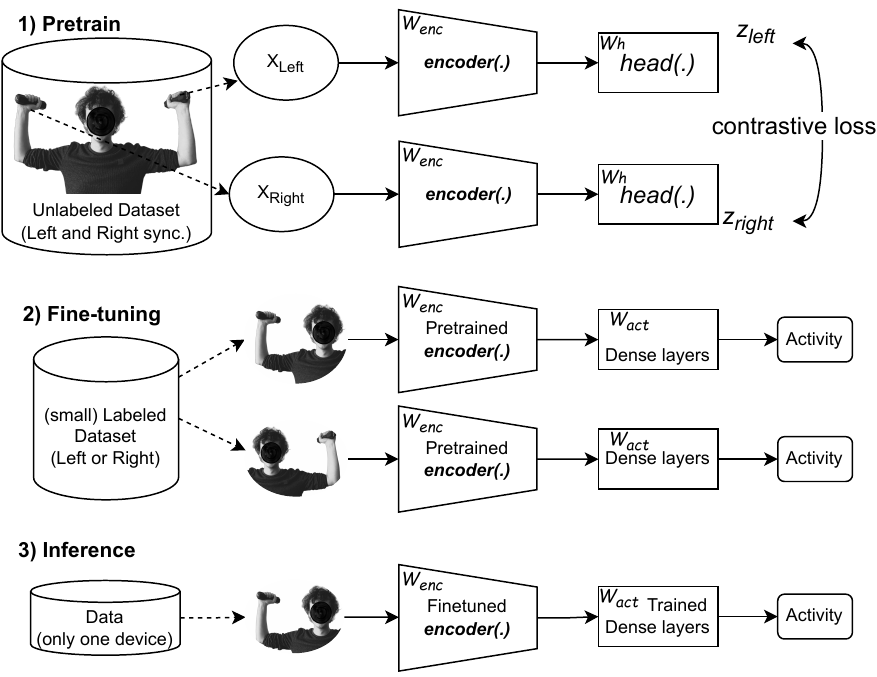}\hspace*{.25\textwidth}%
  \caption{Shows the proposed training settings. When pretraining, we use an encoder and head networks that learn to maximise the agreement of the left and right sensors using unlabelled data. The contrastive loss is utilised, in which the positive samples are left and right sensor data at time $t$ and negatives are taken from the same batch at time $t \prime \neq t$. In step 1, the pretraining process takes pairs of time-synchronised data and trains the encoder in a contrastive way without any label. Next step, the encoder is finetuned using either left or right labelled data for classification. Using both left and right data produces better results. For the inference, the model needs only one device input for prediction.}
  \label{fig:pipeline}
\end{figure}

Machine learning algorithms are improving rapidly, but annotating training data remains a bottleneck for many applications. To overcome this problem, a current area of research focuses on learning effective sensor representations without human-annotated labels for human activity recognition \cite{assessing_har, multitask, selfhar, cpc, enhanced_cpc, reconstruction, reconstruction_gile, masked_reconstruction, simclr, contrastive_learning_har}. These methods follow the pretrain-then-finetune learning schema. In general, the encoder is pretrained in a self-supervised (SSL) way, and then the whole classifier is finetuned with a few labels.
In this work, in the pretraining step (Step 1 in Figure \ref{fig:pipeline}), the network learns from SSL, namely the contrastive prediction task. The contrastive prediction task allows us to train the model without annotations. The resulting model is able to capture relevant characteristics of the input data and is beneficial for the downstream task \cite{assessing_har}, as mentioned in Step 2 of Figure \ref{fig:pipeline}. In the second step in Figure \ref{fig:pipeline}, the downstream task, the model needs a small amount of annotated data to train the classification model.

The success of those methods is already known in other fields such as computer vision \cite{cpcv1_cv, cpcv2_cv, multitask_cv, simclr_cv}, natural language processing \cite{multitask_nlp}, and audio \cite{multitask_audio}. Lastly, similar methods have been proposed for human activity recognition, including contrastive predictive coding \cite{cpc}, multitask \cite{multitask}, and SimCLR \cite{simclr}. The last has a Siamese schema and is based on contrastive learning. Contrastive learning trains a model to match different views of instances created from the same signal by contrasting them. The positive pair is obtained by the transformation of the input window, while the negative pair is formed by the remaining input windows of the same minibatch.

However, specific datasets may not benefit from some SSL methods. For example, in MM-Fit\cite{mmfit}, applying the reconstruction  (autoencoder) as a pretext task did not lead to better classification results. David et al. \cite{mmfit} proposed to learn from different five sensor modalities (fuse inertial sensor data and 3D pose) to be able to deal with the high intra-class variability found in HAR datasets. Pretrain-then-transfer is a strategy utilised in SSL in order to achieve better general representations \cite{assessing_har}, but the design or choice of suitable SSL settings is not trivial \cite{assessing_har, unsupervised_learning}. Therefore, we contribute to the field by proposing a new SSL method that takes advantage of symmetry in activities by using contrastive learning between multiple sensors to tackle issues like (1) limited labelled data or (2) performance on datasets with left-right arm/leg synchronous activity.

Another advantage of contrastive learning is that it can save computational resources. For example, we do not need the preprocessing of raw data as done in \cite{preproc_hand,low_pass}. In addition, other methods \cite{low_pass} utilised handcrafted features, while our contrastive learning-based approach automatically creates useful features.

This work is encouraged by the success of pretraining techniques \cite{assessing_har, collossl, learning_from_the_best} and the necessity of reducing the amount of annotated data for training. We then explore the effectiveness of contrastive learning methods while reducing labelled data requirements\cite{assessing_har}. Our method contrastively matches two different sensor sources (left and right wrist or leg-worn IMUs). The training process requires only one pair of accelerometers. This work further validates the modality-agnostic character of contrastive learning and its potential for generalisation \cite{simclr}.
%\textcolor{red}{contrastive prediction tasks and SSL instead of pretext task}
\section{Related Work}
\label{sec:relatedWork}

% unsupervised and semi-supervised pretraining in other domain (computer vision)
%The success of unsupervised and self-supervised learning has been recorded from different methods and datasets in different domains of computer vision \cite{cpcv1_cv, cpcv2_cv, simclr_cv, byol_cv, multitask_cv}, of natural language processing \cite{multitask_nlp}, etc. Among the pretraining methods, contrastive learning plays an important role for extracting usefully representation.

%\subsection{Contrastive learning}
The work, Contrastive Language-Image Pre-Training (CLIP) \cite{clip}, proposed a technique that aligns pairs of images and texts by projecting them into the joint representation space of CLIP. Moon et al. \cite{imu2clip} used CLIP for aligning Inertial Measurement Unit (IMU) motion sensor recordings with video and text by projecting them into the joint representation space of CLIP. In addition, contrastive learning allows research to create general-purpose training tools such as BYOL \cite{byol_cv, byol_simclr}, SimCLR \cite{simclr_cv, simclr, byol_simclr}, and  contrastive predictive coding (CPC) \cite{cpc, cpcv1_cv, enhanced_cpc, cpcv2_cv}, just to name a few.

% unsupervised and semi-supervised pretraining for HAR
%\subsection{Contrastive learing for HAR}
Contrastive Predictive Coding \cite{cpc, enhanced_cpc} results from the combination of contrastive learning, temporal architecture networks and future timestep prediction. The positive pair is made of long-term temporal network output (context) together with future feature timesteps in the same window, while the negative input of contrastive learning is the samples from other windows of the same batch.

SimCLR \cite{simclr, byol_simclr, contrastive_learning_har} learns general representations from a dataset where a model is trained to match different view instances created from the same signal by contrasting them. The positive pair is obtained by the transformation of the input window, while the negative pair is formed by the rest of the input windows of the same batch. Then, the contrastive loss pulls the positive pair together while pulling the negative pair far away from each other in the representation space.

BYOL \cite{byol_simclr, contrastive_learning_har} relies on two neural networks called online and target networks, that interact and learn from each other. The online network is trained to predict the target network representation of an augmented view of an image. The network is trained to minimise the mean-squared error loss between the online network’s prediction and the target representations. Only the online network is optimised via the loss, and the target network is updated via a slow-moving average of the online network.

Contrastive learning enables the learning and transfer of knowledge from one device to another \cite{collossl, learning_from_the_best}. Previous work has focused on an arbitrary device at an arbitrary position without constraints. The success of our work lies partly in adding constraints for the position of devices, “body symmetry”. Therefore, we focus on devices worn on the left and right sides of the body, for example, the legs or arms.

ColloSSL \cite{collossl} proposes learning useful features from multiple devices by using multi-view contrastive loss. Some of the key innovations in this work are the selection of positive and negative pairs and contrastive sampling techniques. The device selection algorithm is designed to increase the likelihood of selecting ‘good’ positive and negative samples. As metrics, they compute the pairwise Maximum Mean Discrepancy (MMD) among the data points. Then, the positive pair is formed by those with the minimum MMD distance, and the rest are part of the negative samples. To compute the loss function, they applied contrastive sampling, whose role is to decide which data samples should be picked from each device for contrastive training.

%\subsection{Multi-task}
%Multitask for HAR\cite{multitask, selfhar} is pretext task defined as assemble of multiple binary classifications. A single classification task detects whether the original signal is transormed or not. The author \cite{multitask} utilise eight transformations including signal noising, scaling, rotation, negation, flipping, permutation, time warping and channel-shuffling. This allows the model to differentiate between core signal characteristics resiponsable for acitvity categories and other sensor behaviors such as noise, device placements, etc.

%Shared representaion? encoder? translator? classifier? loss?
Closest to our work is Learning from the Best \cite{learning_from_the_best}, a contrastive-based method that transfers useful information from a source sensor to a target one by training a joint representation. While the objective of both works is similar --using synchronised data of two sensors to obtain a better representation --there are several differences. For example, \cite{learning_from_the_best} uses two separate encoders, one for each sensor, while in this work both sensors share the same one. Therefore, we also do not use translator networks to bring one representation to the other. This means that in this work, contrastive learning is applied between the encoded representations of both sensors instead of between the data and its translated version. Another difference is our contrastive loss function: we use the modified NT-Xent\cite{simclr} loss, while previous work utilised the infoNCE loss. Moreover, our networks are simpler, using only 1D-convolution and fully connected layers instead of transformers combined with CNNs.
While we do not compare those two methods in this work, we aim to show that even a simpler version of the contrastive learning scheme can work when taking advantage of symmetries in activities. We hope that this can contribute to position-independent HAR, as our learned model is valid for both devices without compromising performance (e.g., left- or right-hand user).
\section{Method}
\label{sec:method}
Left-Right contrastive learning was inspired by the successful contrastive learning algorithms in computer vision and HAR, such as SimCLR \cite{simclr_cv, simclr}, BYOL \cite{byol_cv, byol_simclr}, learning from the best\cite{learning_from_the_best} algorithms, etc. The proposed method learns useful representations by maximising agreement between left- and right-worn IMU data via a contrastive loss in the latent space. In contrast to \cite{learning_from_the_best}, our method uses one encoder for both sensors and has no translator network. Our method has the advantage of a simpler model, while \cite{learning_from_the_best} has the ability to train with data from any pair of devices. The training procedure illustrated in Figure \ref{fig:pipeline}, presents the following four major components:
\begin{enumerate}[1.]
    \item The \textbf{left and right sensor data} synchronised within time allows building the input $x$ of training without the need for data augmentation. The stochastic data augmentations used in other contrastive methods reflect an approximation of sensor properties. Using directly raw sensor data as views allows learning the sensor properties directly instead of through the view's approximation by transformations.
    \item An \textbf{encoder network} $encoder(.)$ that predicts the representation in the latent space from the input data. The training pipeline has no restriction on architecture; a neural network of choice can be used. For example, we use the encoder from \cite{simclr,multitask} which has three (1D) convolutional layers.
    \item A \textbf{project head} that takes the representation $h=encoder(x)$ and projects it into contrastive latent space $z=head(h)$, s. Figure \ref{fig:pipeline}.
    \item A \textbf{contrastive loss} function that defines agreements within positive and negative samples in the same minibatch. 
\end{enumerate}

The proposed method can be trained using different tools of siamese architecture and contrastive loss with simple modifications such as BYOL \cite{byol_cv}, SimSiam \cite{simsiam}, or SimCLR \cite{simclr_cv}. We chose to modify the contrastive learning framework SimCLR so that we use left and right sensor data instead of a double transformation.
As in \cite{simclr_cv, simclr, byol_simclr}, we randomly sample a minibatch of $N$ examples and define the contrastive loss with pairs of left $x^{left}$ and right $x^{right}$ sensor data derived from the minibatch. Further details can be found in Algorithm \ref{alg:algorithm}.

\begin{algorithm}[hbt!]
\caption{Left-Right contrastive learning.} \label{alg:algorithm}
    \begin{algorithmic}
    \Require{minibatch of size N, temperature constant $\tau$}
    \Ensure{$\mathcal{L}$ loss}
    \For{sampled minibatch $\{x_k\}_{k=1}^N$}                    
        \For{all $k \in \{1...N\}$}
            \State $z_{2k-1} = head(encoder(x_k^{left}))$
            \State $z_{2k} = head(encoder(x_k^{right}))$
        \EndFor
        \For{all $i \in \{1...2N\}$ and $j \in \{1...2N\}$}
            \State $sim_{i,j} \gets$ pairwise similarity between $z_i$ and $z_j$
        \EndFor
        \State $l(i, j) \gets - \log \frac{\exp(sim_{i, j}/\tau)}
                                      {\sum_{k=1}^{2N} \mathbb{1}_{[k \neq i]} \exp(sim_{i, j}/\tau)}$
                                      \Comment{defines loss for a positive pair ($i$ and $j$)}
        \State $\mathcal{L} = \frac{1}{2N} \sum_{k=1}^N[l(2k-1, 2k) + l(2k, 2k-1)]$
    \EndFor
\end{algorithmic}
\end{algorithm}

Note that the output of this network is normalised to lie on the unit of the hypersphere, which allows using the inner product as similarity metrics for contrastive learning \cite{simclr_cv, cpcv1_cv}. The projection head is only used while training and does not participate in inference settings. In addition, the pipeline applies cosine similarity $sim(.)$ as a similarity function. The remaining details are in experiments section. %will be published in the code.
\section{Experiment}
\label{sec:expriment}

In this section, we present the datasets used in our experiments. First, we give a short description of our training pipeline and evaluation metrics. Finally, the results are presented.

\subsection{Evaluation}
We start with an overview of the two datasets used, MM-Fit and Opportunity. Afterwards, the details of the training procedure are presented along with the network's architecture. In addition, the validation procedure is explained.
\begin{figure}[ht]
  \centering
  \includegraphics[page=1,width=0.9\textwidth]{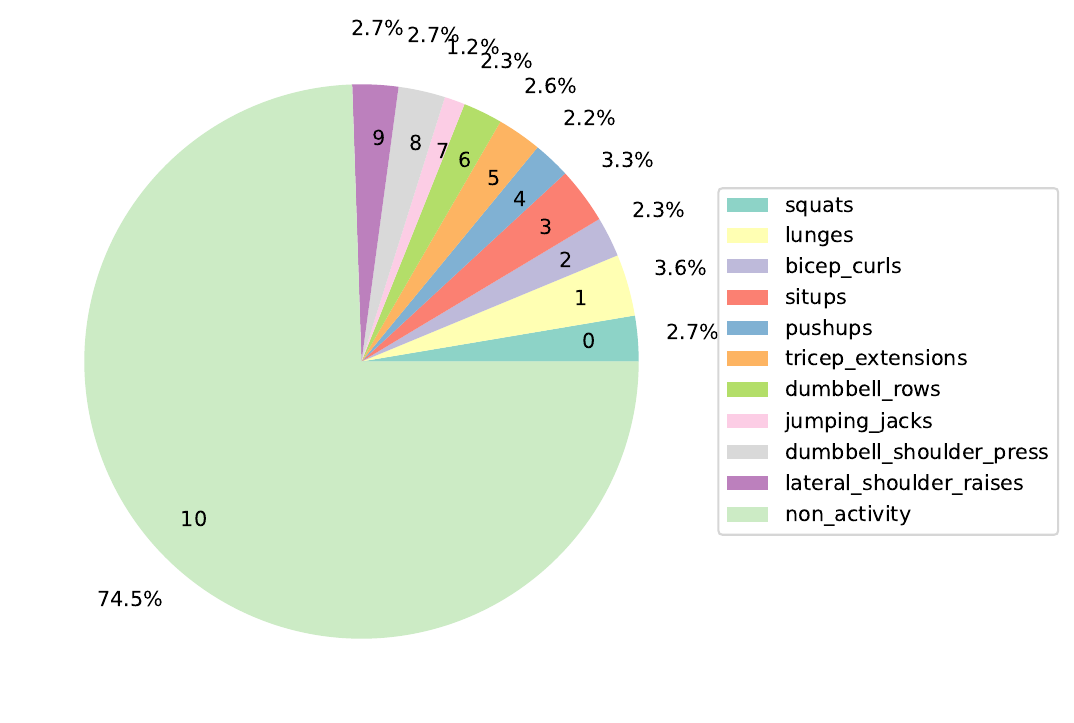}\hspace*{.25\textwidth}%
  \caption{Shows MM-Fit class distribution based on windows of two seconds with 50\% overlap. The preprocessed dataset has 29175 samples.}
  \label{fig:mmfit_classes}
\end{figure}

\textbf{MM-Fit Dataset.}
An open-source dataset, MM-Fit \cite{mmfit}, is used in our experiments. The dataset is collected from different body-worn inertial sensors and cameras (not worn), which are time-synchronised. There are four types of devices, including the \textit{Orbbec Astra Pro} camera, the \textit{eSense (Nokia Bell-Labs)} earbud, the two \textit{Mobvoi TicWatch Pro} smartwatches, and two smartphones: the \textit{Samsung S7} and \textit{Huawei P20}. The subject wore two smartwatches, one on each wrist, an earbud in the left ear, and smartphones in the left and right trouser pockets. The smartwatches recorded the acceleration and angular velocity at $100$ Hz. In this work, we use the accelerometers from the two smartwatches. We apply windows of two seconds with a step of  one second. The MM-Fit contains ten activities and 'no activity' classes; among them, we cite: squats, lunges (with dumbbells), bicep curls (alternating arms), sit-ups, push-ups, sitting overhead dumbbell triceps extensions, standing dumbbell rows, jumping jacks, sitting dumbbell shoulder presses, and dumbbell lateral shoulder raises. In addition, annotation includes repetition counting across multiple devices. For this dataset, we take into account all the described classes. Figure \ref{fig:mmfit_classes} provides a short overview of class distribution based on a window of two seconds with 50\% overlap. We notice that the data are not well balanced, with the null class dominating the dataset.

\textbf{Opportunity Dataset.}
The Opportunity dataset \cite{opportunity_collecting} recorded sensor data for 6 hours from four subjects in a daily living (ADL) scenario. The sensors were either integrated into the environment, or in objects, or on the body. The activities in the experiments are grouped as locomotion or gestures. This work focuses on locomotion activities where there are four classes, namely stand, walk, sit, and lie, as in \cite{opportunity_benchmark,learning_from_the_best}. This dataset has more than one label type, for example, mode locomotion and mode gestures \cite{opportunity_benchmark}. 
We evaluate the locomotion classes. In terms of the sensor settings, we choose acceleration data from the left and right lower arms (RLA and LLA), each having a sampling rate of $30$ Hz.
The null class was excluded from our experiments as, unlike in the case of MM-Fit, here we have many static classes and there is too much ambiguity between them and the null class as we are using only a single accelerometer for inference.
This is also the reason we have not evaluated our method based on gestures. In other words, we follow this evaluation procedure because we focus on HAR from a single sensor, and it can be challenging to predict all classes using only a single acceleration data. For example, some activities (e.g., open, close, etc.) from the mode gesture are exclusively performed by one dominant hand \cite{Opportunity_pp}. Handling those classes is future work and would have to include more than one sensor as source data, as with a single one for inference, the signals to the corresponding activity would be missing.

\textbf{Metrics.}
Three common metrics are used to study the performance of the models: macro F1, weighted F1, and accuracy. We run every experiment ten times and report the mean and standard deviation. As a baseline, we trained one model using left and right accelerations and then compute the metrics separately on left and right devices. Later, we noted 'both' to mention when left and right sensor data are used together; otherwise, it was specified whether we utilise data from the left or right arm.

\begin{figure}[ht]
  \centering
  \includegraphics[page=1,width=0.9\textwidth]{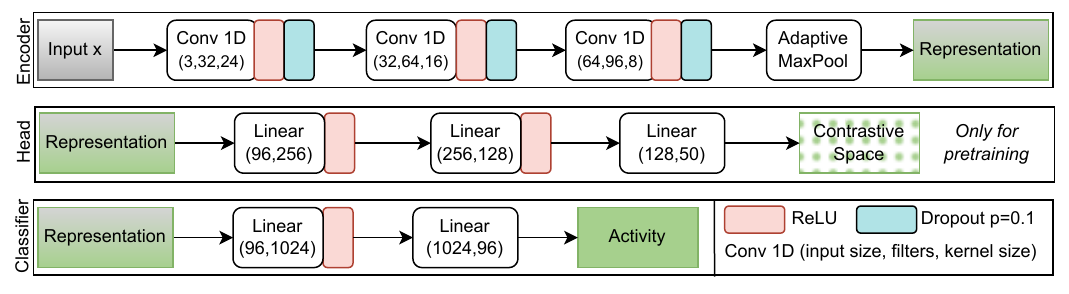}\hspace*{.15\textwidth}%
  \caption{Shows the architecture of the network. The adaptive max-pooling is applied over time. A dropout rate of $10\%$ is used. The model is simple; the encoder and classifier together comprise $146\:thousand$ parameters.}
  % The architecture of the networks, encoder and classifier.
  \label{fig:architecture}
\end{figure}
\textbf{Training procedure.}
Regarding the architecture of the networks (Figure \ref{fig:architecture}), for pretraining, we need an encoder and a prediction head.
The encoder has three 1D-convolutional layers with kernel of $24, 16, 8$, and $32, 64, 96$ filters, respectively. Every layer is followed by a relu activation function and a dropout of $0.1\%$. We applied global maximum pooling to the feature map in the third layer. In this step, we used the contrastive prediction head, which consists of three dense layers with $256, 128$, and $S$ output sizes, respectively. The contrastive loss was then optimised using SGD  with a cosine decay on learning rate of $0.004$, a temperature of $0.05$, and a batch size of $64$ over $200$ epochs. Note that SimCLR was pretrained using rotation as the transformation, the same as in the code published with the paper \cite{simclr}.

For classification, we used the encoder with previously trained weights, followed by a classifier module.
We chose a simple classifier with two fully connected layers with $1024$ and class numbers respectively. The model is finetuned based on cross-entropy loss with the Adam optimiser, with a learning rate of $0.0001$ for $50$ epochs.

All our results refer to an unseen test set. As it is done in \cite{mmfit}, the MM-Fit dataset is split according to the participants data: train $(1, 2, 3, 4, 6, 7, 8, 16, 17, 18)$, validation $(14, 15, 19)$, test set $(9, 10, 11)$, and unseen test set $(0, 5, 12, 13, 20)$.
%We evaluated best model using unseen test set and rapport the result.
For Opportunity, data are preprocessed similar to \cite{DeepConvLSTMOpportunity}, where we train the model on the data of all ADL, and drill sessions for the first subject and on ADL1, ADL2, ADL3 and drill sessions for Subjects 2 and 3. We report classification performance on a testing set composed of ADL4 and ADL5 for Subjects 2 and 3.

In order to evaluate the quality of learned representations, we used reduced labelled data for finetuning, namely $1, 5, 10, 50$, and $100$ samples per class.

The training follows the protocol in Figure \ref{fig:pipeline}, which implements the schema of pretrain-then-finetune. In the first step, we train the encoder in our proposed self-supervised learning manner.
We trained with different hyper-parameters as done in \cite{enhanced_cpc}, then we chose the best encoder based on validation.

In the second step, the classification task, we used early stopping with patience of 5 and kept the best classification model. As in \cite{multitask, simclr}, all layers of the base encoder are frozen except for the last one, and a two-layer MLP (Multi-Layer Perceptron) classifier is added and finetuned.

\subsection{Results}

In Table \ref{table:performance}, we can see the performance of our method when compared to the baseline (supervised training) and another SSL method (SimCLR). There we can see that for the MM-Fit dataset, our method presented around 8\% macro f1 improvement compared to the baseline and even bigger gains when compared to SimCLR, which did not beat the baseline. This shows the potential of our approach, which, unlike SimCLR, does not rely on data transformations. Since MM-Fit includes symmetric motions, it is likely our method's improvement comes directly from the lack of data transformations. One could think that having access to double the data is the source of our improvement, but our method also beats a supervised approach trained using data from both sensors as examples, which points to our contrastive training itself being beneficial.

\begin{figure}[ht!]
    \centering
    \includegraphics[width=\textwidth]{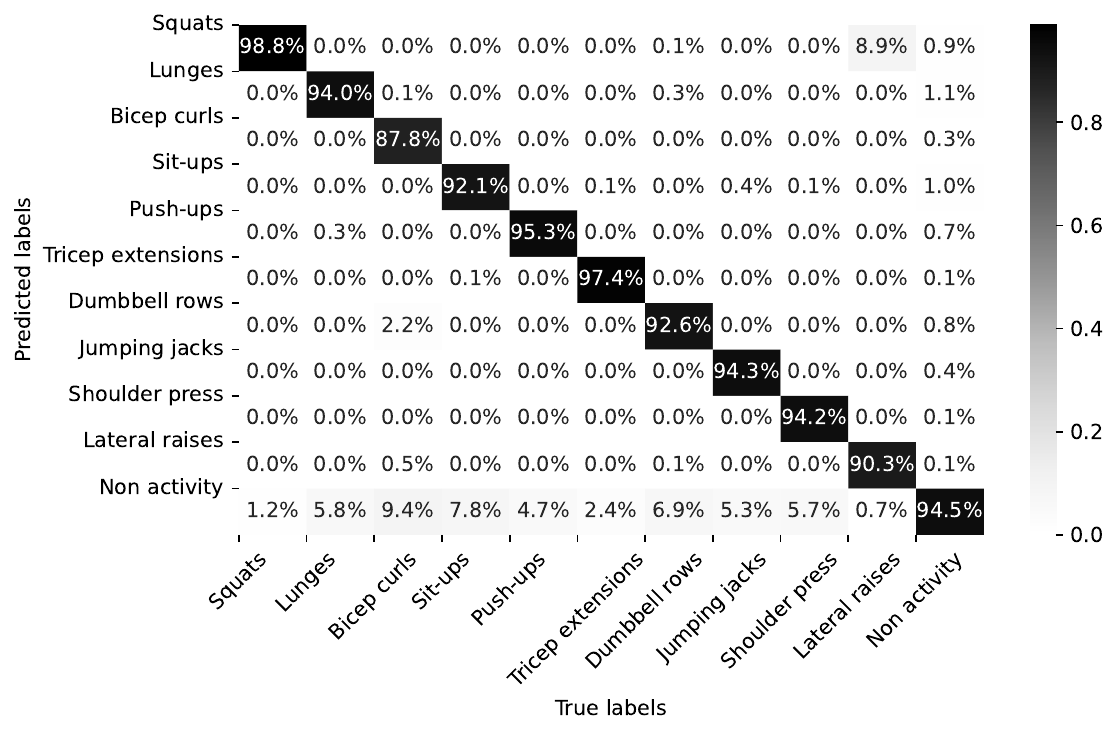}
    \caption[Performance presented as confusion matrix]{Shows confusion matrix on MM-Fit for unseen subjects with acceleration data from right smartwatch.}
    \label{fig:confusion_matrix}
\end{figure}

Since MM-Fit is an unbalanced dataset, as seen in Figure \ref{fig:mmfit_classes}, it is relevant to look at the confusion matrix in Figure \ref{fig:confusion_matrix} to show our performance in the most and least frequent classes. For example, in addition, the bicep curls exercise presented the lowest performance with $87.8\%$ accuracy, which can be expected as it is the least symmetric activity. The highest performance was observed in the squats exercise, with $98.8\%$ accuracy. The non-activity class achieved an accuracy score of $94.5\%$, while the remaining activities had a mean accuracy of $93.7\%$. This shows that the proposed method is able to handle imbalanced data.

\begin{table}
    \centering
    \begin{tabular}{c c c c c c} 
    Method                                  & Training  & Inference & F1 Weighted  & F1 Macro      & Accuracy   \\
    \toprule
    \multirow{2}{*}{Baseline}               & Left      & Left      &90.26±0.67& 74.77±1.64    & 91.00±0.70 \\
                                            & Right     & Right     &90.97±0.68& 76.52±2.26    & 91.72±0.57 \\
    %\midrule
                                            & \multirow{2}{*}{Both} &
                                                          Left      &89.82±0.54& 72.28±1.42    & 90.48±0.62 \\
                                            &           & Right     &89.76±0.65& 72.29±2.50    & 90.77±0.54 \\
    \midrule
    \multirow{2}{*}{SimCLR\cite{simclr}}    & Both      & Left      &89.93±0.54& 69.58±1.76    & 90.69±0.43 \\
                                            & Both      & Right     &89.91±0.75& 69.90±2.49    & 90.69±0.58 \\
    \midrule
    \multirow{2}{*}{MM-Fit\cite{mmfit}}     &           & Left      &&               & 90.74      \\
                                            &           & Right     &&               & 92.72      \\
    \midrule
    \multirow{2}{*}{Ours}                   & \multirow{2}{*}{SSL+Both}   &
                                                          Left      &93.19±0.20& 82.99±0.77    & 93.57±0.16 \\
                                            &           & Right     &93.93±0.53& \textbf{85.28±1.86}    & \textbf{94.31±0.42} \\
    %\bottomrule
    \end{tabular}
    \caption{Classification (macro f1 and accuracy) results for the proposed method with comparison to baseline, MM-Fit (result reported from paper \cite{mmfit}) and SimCLR.}
    \label{table:performance}
\end{table}

The results on the Opportunity dataset can be read from Table \ref{tab:performance_opportunity}, where results from baseline, SimCLR, and the proposed method are compared. All SSL methods show promising improvements at all metrics of more than $25\%$. 
%Especially the proposed method performed better than baseline and other SSL method (SimCLR) with macro f1 of $73.83\%$ and $62.74\%$ on left and right accelerometer respectively.

In this case, our method still outperforms SimCLR, but now by less than 2\% of the macro F1 score. More specifically, pretraining enhanced the results from $0.39$ to $0.73$ and $0.74$ macro f1 for SimCLR and the proposed method, respectively (left accelerometer).

\begin{figure}[t!]
    \centering
    \subfigure{\includegraphics[width=0.48\textwidth]{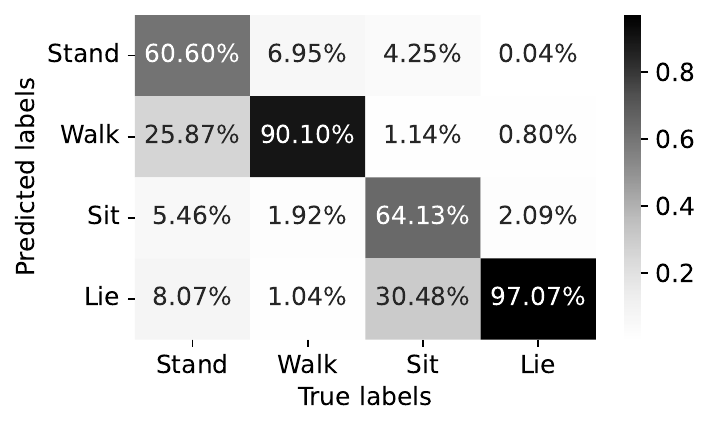}}
    \subfigure{\includegraphics[width=0.48\textwidth]{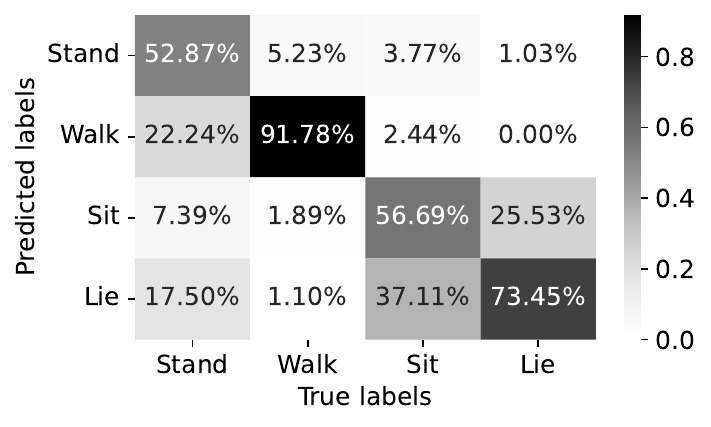}}
    \caption{Shows the confusion matrix on Opportunity dataset with proposed method. Model is tested with data from left hand (left figure) and right hand (right figure).}
    \label{fig:confusion_matrix_opp}
\end{figure}

We hypothesise that our gains are smaller when compared to SimCLR due to the locomotion activities being less symmetric. Figure \ref{fig:confusion_matrix_opp} shows that while lying and walking activities are well recognised in the Opportunity dataset, standing and sitting are the most challenging activities for the model. 

\begin{table}
    \centering
    \begin{tabular}{c c c c c c} 
    Method                                  & Training                    & Inference & F1 Weighted & F1 Macro   & Accuracy   \\
    \toprule    
    \multirow{4}{*}{Baseline}               & Left                        & Left      &49.30±3.97& 38.51±3.35 & 60.78±1.84 \\
                                            & Right                       & Right     &43.40±3.52& 34.39±6.74 & 54.60±3.29 \\
                                            & \multirow{2}{*}{Both}       & Left      &48.36±1.03& 37,73±0.86 & 60.32±1.34 \\
                                            &                             & Right     &37.35±4.42& 27.94±4.06 & 48.95±3.82 \\
    \midrule
    \multirow{2}{*}{SimCLR\cite{simclr}}    & \multirow{2}{*}{Both}       & Left      &76.17±0.85& 72.56±2.00 & 77.02±0.80 \\
                                            &                             & Right     &71.08±0.81& 60.57±1.15 & 72.24±1.30 \\
    \midrule
    \multirow{2}{*}{Ours}                   & \multirow{2}{*}{SSL+Both} & Left      &77.53±0.93& \textbf{73.84±1.63} & \textbf{78.30±0.89} \\
                                            &                             & Right     &72.99±0.93& 62.74±1.68 & 74.27±1.21
    \end{tabular}
\caption{Classification results for the proposed method with comparison to baseline and SimCLR on Opportunity dataset \cite{opportunity_benchmark, opportunity_collecting}.}
\label{tab:performance_opportunity}
\end{table}

\subsection{Effect of a reduced training set}
\begin{figure}[ht!]
    \centering
    \includegraphics[width=250pt]{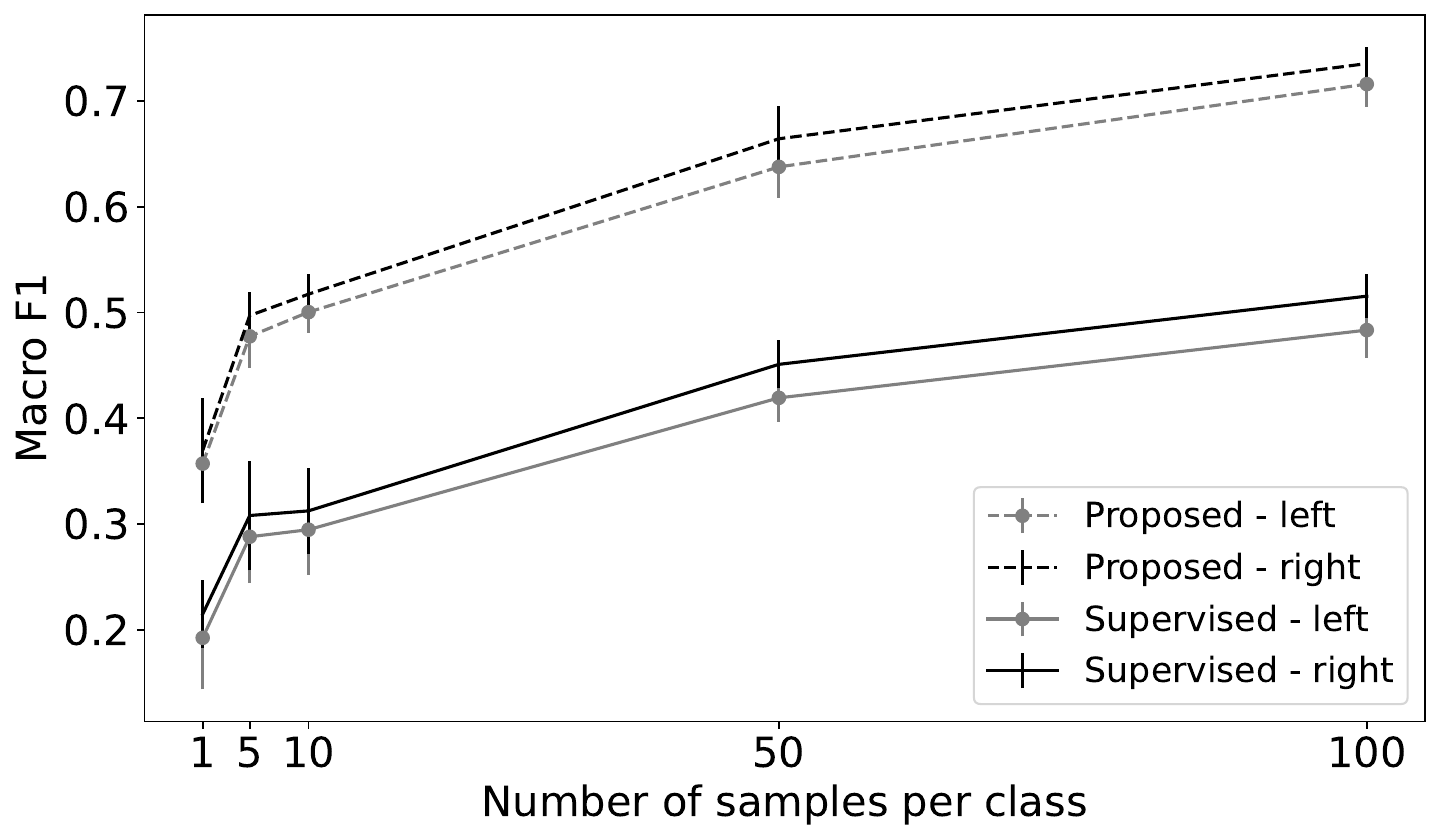}
    %%\centerline{\epsfig{/Chapters/chapter1/figures/cat.eps,width=.8\textheight,height=.4\textwidth}}
    \caption[Performance in MM-Fit with reduced fine tuning training set]{Indicates performance of pretrained model while finetuning with reduced training set in MM-Fit (proposed settings). Supervised refers to the training from scratch (our baseline).}
    \label{fig:reduced_train_sest}
\end{figure}

The Figure \ref{fig:reduced_train_sest} shows results for an experiment in which we vary the number of labelled data available in the MM-Fit dataset. Here, we can see that using only a small amount of labelled data (100 samples per class), we can achieve a macro f1 score of 0.7 (dashed lines) with more than $20\%$ improvement compared to training from scratch with the same amount of examples.

\begin{figure}[t!]
    \centering
    \subfigure{\includegraphics[width=0.52\textwidth]{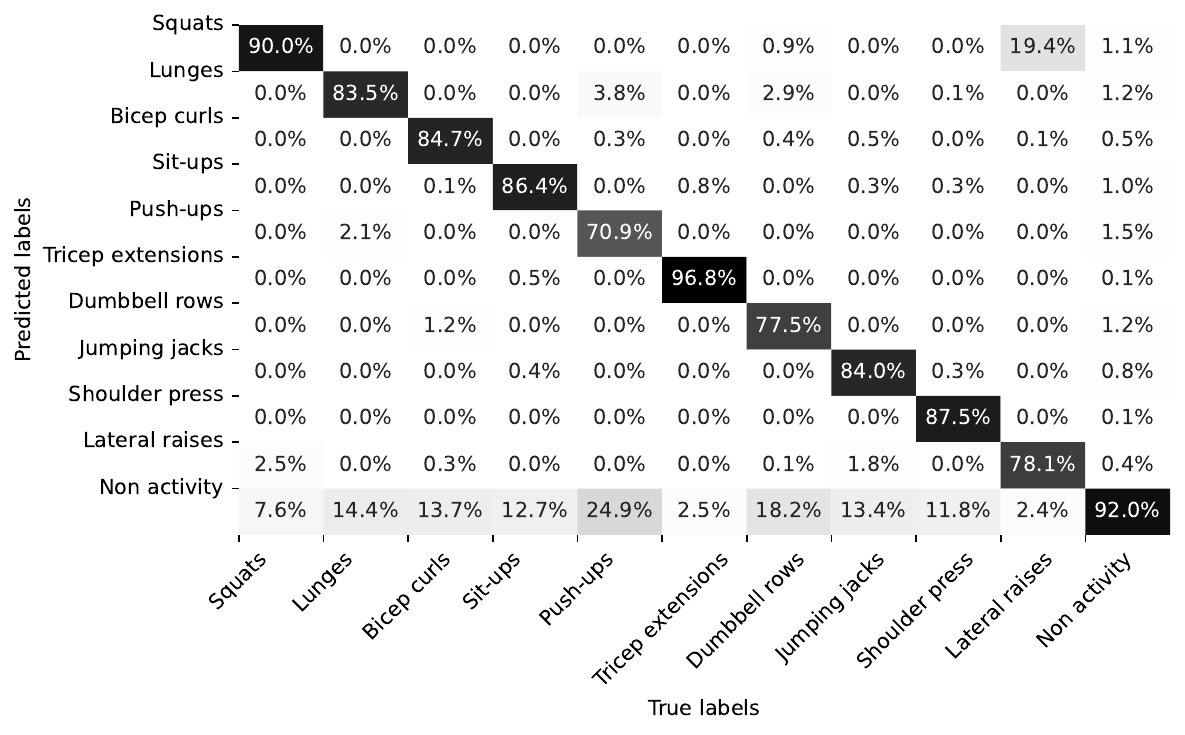}}
    \subfigure{\includegraphics[width=0.47\textwidth]{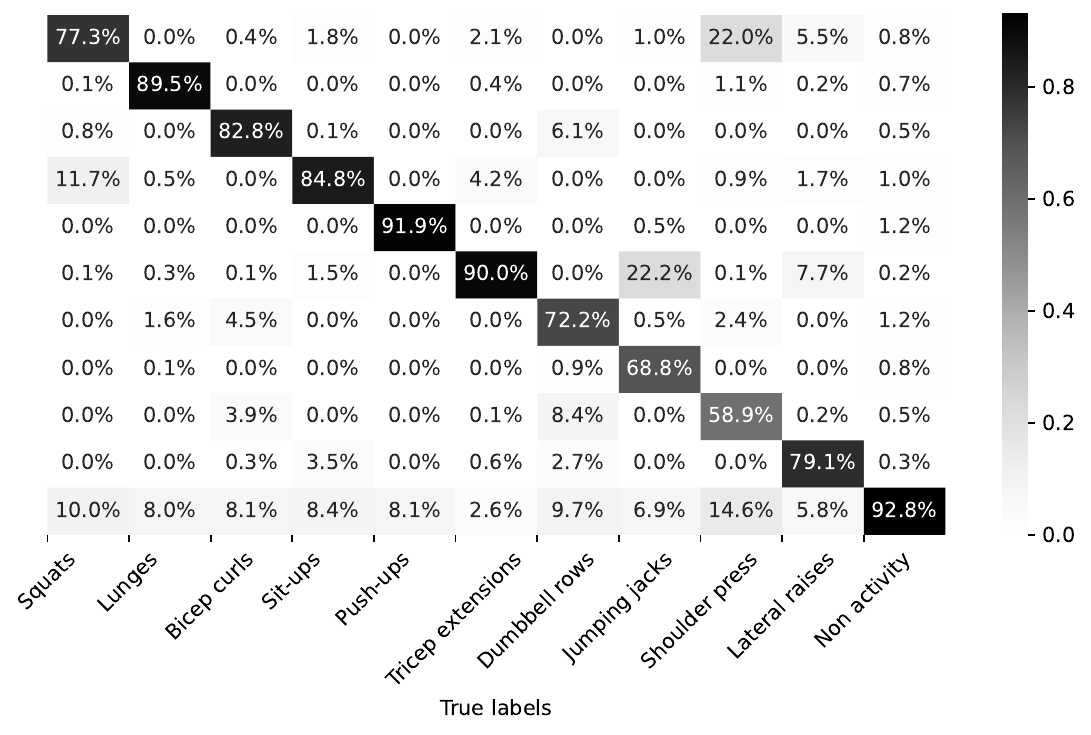}}
    \caption{Shows confusion matrix (left arm) on MM-Fit dataset for baseline (left) and SimCLR (right) methods.}
    \label{fig:confusion_matrix_mmfit}
\end{figure}

%\begin{figure}[h!]
%    \begin{center}
%    \subfigure[\label{f8a}]{\includegraphics[width=5.4cm,height=4.5cm]{Authors/figures/confusion_matrix_left_wocb.pdf}}
%    \subfigure[\label{f8b}]{\includegraphics[width=5cm,height=4.5cm]{Authors/figures/confusion_matrix_right_woy.pdf}}
%    \end{center}
%    \caption[Performance presented as confusion matrix]{\textcolor{red}{Shows confusion matrix with ... activities in MMFit.}}
%    \label{fig:confusion_matrix}
%\end{figure}
\begin{figure}[t!]
    \centering
    \subfigure{\includegraphics[width=0.48\textwidth]{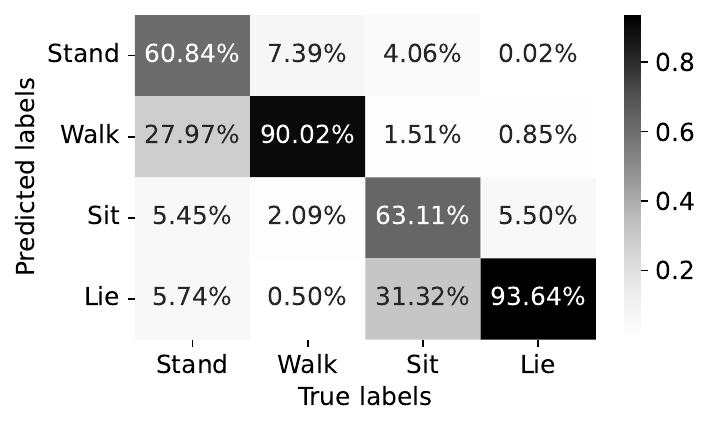}}
    \subfigure{\includegraphics[width=0.48\textwidth]{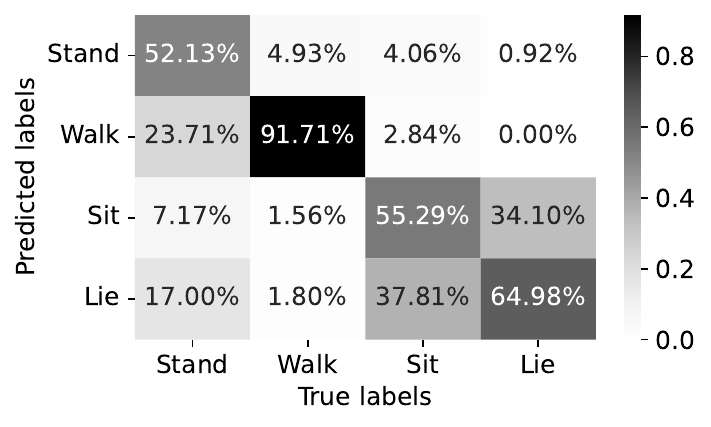}}
    \caption{Shows the confusion matrix on Opportunity dataset with SimCLR. The model is tested with data from left hand (left figure) and right hand (right figure).}
    \label{fig:confusion_matrix_opp_simclr}
\end{figure}

\subsection{Contribution of batch size to contrastive loss}
\begin{figure}[ht!]
    \begin{center}
    \subfigure[\label{la}]{\includegraphics[width=0.475\textwidth]{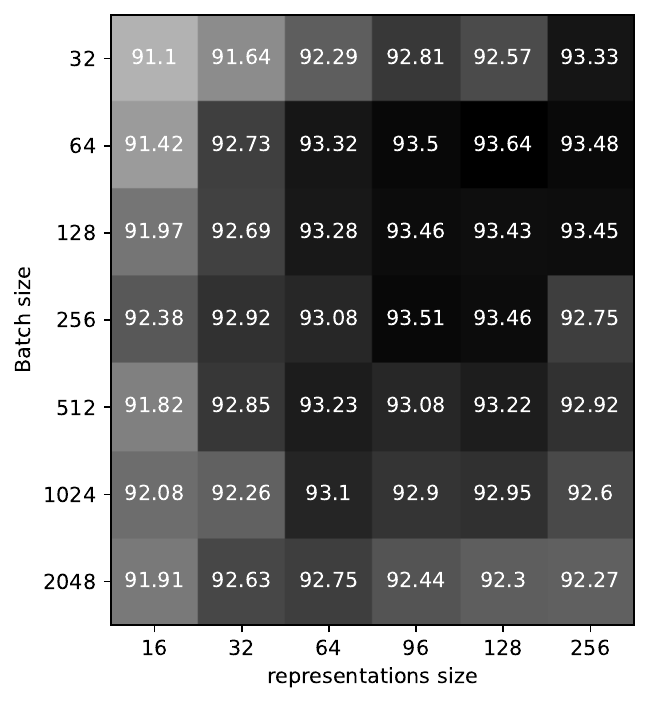}}
    \subfigure[\label{lb}]{\includegraphics[width=0.515\textwidth]{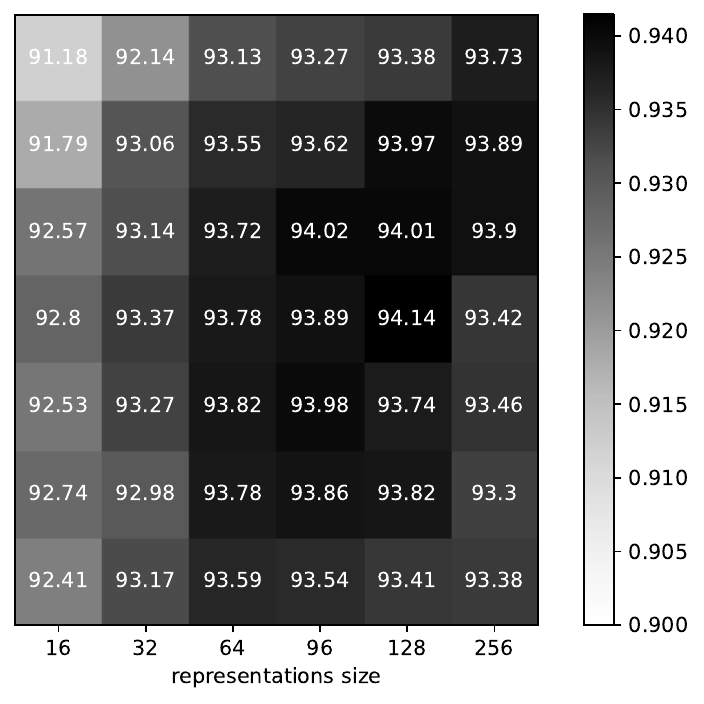}}
    \end{center}
    \caption[The Figure displays how batch size and latent space size affect performance]{Displays the effect (accuracy in percentage) of batch size and latent space size. Figure (a) and Figure (b) report on left and right worn smartwatch, respectively.}
    \label{fig:batch_repres}
\end{figure}

We also did an analysis of the contribution of batch size to contrastive loss. We could not observe the positive relationship between batch size and performance, in contrast to \cite{simclr_cv}, where a large minibatch performs better. Figure \ref{fig:batch_repres} shows the contribution of batch size and latent space size to performance. As the minibatch size increases, the performance does not necessarily become better. The same behaviour is observed while comparing latent space size and accuracy.

\subsection{Discussion and Limitations.}
The proposed method outperforms the baseline and SimCLR in all tasks. It achieves $84.03\%$ accuracy with an improvement of $35\%$ compared with training from scratch while having only 100 labelled samples (7,76\% of the training set on MM-Fit). In addition, we observed improvements on MM-Fit and Opportunity if the entire labelled data (for trains) were available. However, the proposed method does not perform significantly better than other SSL on datasets, e.g., Opportunity, whose activities do not synchronously involve both arms or no arms, e.g., standing or sitting. Details can be visualised by comparing Figure \ref{fig:confusion_matrix_opp} and Figure \ref{fig:confusion_matrix_opp_simclr}. The reason behind this may lie in the fact that the motion signals captured from the arms are perturbed by (1) independent arm movement or (2) signal propagation through different joints \cite{learning_from_the_best}. But there are some open questions, for example, about the performance when the dataset contains a large number of locomotion activities. Moreover, it is not always possible (or easy) to collect left-right synchronous data for various reasons, e.g., running, walking, etc.

Looking at the improvement done by the proposed method, we could observe that some classes improved only marginally from baseline. For example, the lowest improvement is observed in tricep extensions activity, where the proposed method has a $0.6\%$ gain in macro F1 from baseline; for details, see Figures \ref{fig:confusion_matrix} and \ref{fig:confusion_matrix_mmfit}. In contrast to lower improvements, SimCLR classifies worse than baseline; for details, see Figure \ref{fig:confusion_matrix_mmfit}. Note that triceps extension in MM-Fit is symmetrical for the left and right arms. Perhaps the proposed method makes use of this synchronous behaviour, but in this case, the improvement is not significant.
\section{Conclusion}
\label{sec:conclusion}

The proposed pretraining technique applies contrastive learning to two different sensor data from left and right wrist-worn devices and hence does not need any transformation. In order to evaluate the performance of the learned representation for HAR, we followed the schema of 'pretrain-then-finetune'. In the results, we observed that left-right contrastive learning draws promising results. In addition, the finetuned model performed better than fully supervised learning from scratch. The results also showed the role of data representation, where the model achieved around a $0.7$ macro f1 score on MM-Fit using only 100 annotated samples per class.

We have observed that our method does not need a large minibatch, in contrast to other contrastive learning frameworks based on transformations. Moreover, it can alleviate data requirements by still showing improvements given a reduced training set.

On the other hand, our proposed method was designed with symmetrical activities in mind. Thus, it requires synchronised data from both sides and may not provide substantial improvements if only one of the limbs is involved in the activity. Therefore, in future work, we will evaluate our methods with more datasets that are less symmetrical in motion.

In many areas, prior knowledge about the activities being performed is known beforehand. Thus, we will investigate how our method can be applied to those activities that fit this proposed schema. We will also investigate how to generalise the method so that other relationships between activities can be included in the contrastive framework.

Moreover, future work will also focus on studying the effects of different modalities and sensor placement, as well as data transformations.
\section{Acknowledgments}

The research reported in this paper was supported by the Carl Zeiss Stiftung, Germany under the Sustainable Embedded AI project (P2021-02-009).

\bibliographystyle{frontmatter/spmpsci.bst}
\bibliography{Authors/bibtex}
%%\bibliographystyle{plain}
%%\bibliography{bibtex}

%%\printindex
%%\cleardoublepage
\end{document}